\documentclass[runningheads]{llncs}
\usepackage[T1]{fontenc}
\usepackage{graphicx}
\usepackage{algorithm} 
\usepackage[noend]{algpseudocode}
\usepackage{xcolor}
\usepackage{amsmath}
\usepackage{multirow}
\usepackage{array}
\usepackage{amssymb}
\algdef{SE}[SUBALG]{Indent}{EndIndent}{}{\algorithmicend\ }%
\algtext*{Indent}
\algtext*{EndIndent}
\usepackage{hyperref}

\begin{document}
\title{Online state vector reduction \\
during model predictive control \\
with gradient-based trajectory optimisation}
\titlerunning{State vector reduction during MPC}
%
\author{David Russell\orcidID{0009-0002-5660-3890} \and
Rafael Papallas\orcidID{0000-0003-3892-1940} \and
Mehmet Dogar\orcidID{0000-0002-6896-5461}}
\authorrunning{D. Russell et al.}
%
\institute{School of Computer Science, University of Leeds, Leeds, LS2 9JT, UK \\
\email{el16dmcr@leeds.ac.uk}
} 
%
\maketitle              
\begin{abstract}
Non-prehensile manipulation in high-dimensional systems is challenging for a variety of reasons. One of the main reasons is the computationally long planning times that come with a large state space. Trajectory optimisation algorithms have proved their utility in a wide variety of tasks, but, like most methods struggle scaling to the high dimensional systems ubiquitous to non-prehensile manipulation in clutter as well as deformable object manipulation. We reason that, during manipulation, different degrees of freedom will become more or less important to the task over time as the system evolves. We leverage this idea to reduce the number of degrees of freedom considered in a trajectory optimisation problem, to reduce planning times. This idea is particularly relevant in the context of model predictive control (MPC) where the cost landscape of the optimisation problem is constantly evolving. We provide simulation results under asynchronous MPC and show our methods are capable of achieving better overall performance due to the decreased policy lag whilst still being able to optimise trajectories effectively.


\end{abstract}

\section{Introduction} \label{Introduction}

\begin{figure}[t] \label{push_mcl_snapshots}
\includegraphics[width=\textwidth]{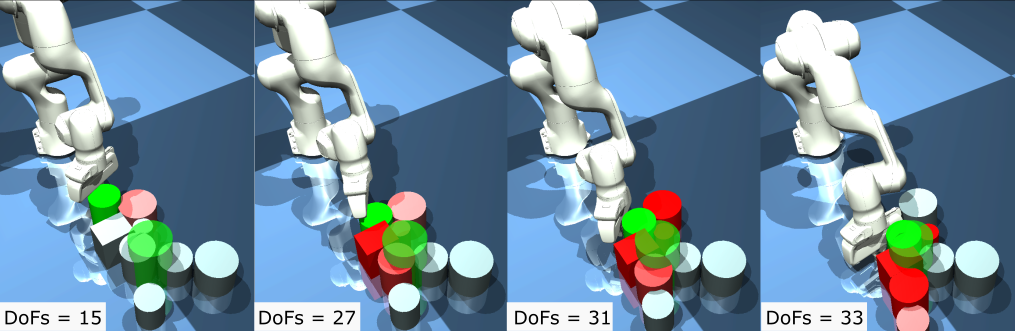}
\caption{A sequence of snapshots showing an example MPC trajectory generated by our method. The task is to push the green cylinder to a goal region (the green transparent cylindrical region) whilst minimally disturbing some clutter objects. The full number of DoFs in this system is 55. Our method identifies the relevant DoFs of this system at different times during execution and performs trajectory optimisation using this reduced state. Objects with stronger shades of red have more DoFs in the state vector at that point during execution: If an object is dark red, all of its six DoFs are considered; if an object is white, none of its six DoFs are considered during trajectory optimisation.}
\end{figure}

Trajectory optimisation algorithms are capable of synthesising complex motion to solve a variety of challenging robotic manipulation tasks \cite{Kitaev_2015,Kurtz_2022_iLQR,selvaggio_2023_manip,Tassa_2012_iLQR,Onol_2020_optimization,papallas_2022_ask}. However, trajectory optimisation methods struggle scaling to high dimensional systems, such as manipulation in cluttered environments or manipulation on deformable objects. The long optimisation times caused by the curse of dimensionality make it difficult to perform closed-loop model predictive control in these high-dimensional systems.

In this paper, we address this problem by trying to reduce the dimensionality of these systems dynamically so that closed-loop model predictive control (MPC) can be achieved. Our key insight is that, in a variety of these high-dimensional tasks, a large number of degrees of freedom (DoFs) are not relevant to the problem at all times during task execution. This intuition makes sense on a fundamental level: Humans do not consider the full physical effects of all DoFs in a system when performing various tasks. When we fold clothes we do not consider in great detail the path each particle in the cloth will take; similarly, when we reach into a cluttered shelf, we are good at identifying which objects matter for our goals and ignoring others.

Fig.~\ref{push_mcl_snapshots} shows a trajectory generated by our method for a non-prehensile manipulation task in clutter. Traditionally, during trajectory optimisation in such a scene, all six DoFs of all movable rigid objects (which contribute twelve state vector elements, the positional and velocity element of each DoF) are considered in the system state, in addition to the robot DoFs. Instead, we propose a method that can identify the relevant DoFs at different times during the task, and perform trajectory optimisation only with the reduced system state.


Reduced order models for efficient planning and control have been used in robotics before.  Locomotion is one such area where reduced order models have found great performance in enabling real time closed-loop control of high-dimensional robots \cite{blickhan_1989_spring,kajita_1991_study}.
These can be pre-defined, such as inverted-pendulum models, or learned models given a task description \cite{Chen_2020_dim_reduction}. While it is possible to take such approaches for locomotion (where the full model is almost always limited to the robot, and the reduced model is a lower-dimensional approximation of the same robot), it is more difficult to take a similar approach to object manipulation tasks as in Fig.~\ref{push_mcl_snapshots}, since the full model can include an arbitrary number, shape, and configuration of objects, which are unknown beforehand. That is why in this work we consider the method of reducing dimensionality \textit{online} during task execution (MPC), given a task instance. Furthermore, the reduced state can be different for different stages of the task, as shown in the different snapshots in the figure. 

It is important to note that we do \textit{not} extract a reduced \textit{dynamics} model of the system; rather, we only extract a reduced state vector. We show that there are significant gains (in terms of optimisation time) to be made with a reduced state vector, specifically when using gradient-based shooting trajectory optimisation methods, e.g., the iterative linear quadratic regulator (iLQR) \cite{Li_2004_iLQR}. Such methods require derivatives of the system dynamics to be calculated, often using computationally expensive \textit{finite differencing} \cite{russell_2023_adaptive}, which benefit significantly from a smaller state vector. Other operations within trajectory optimisation, such as the backwards propagation of the value function, also benefit from reduced state sizes. Therefore, in this paper, we also present our modifications to the iLQR method, which enable us to use it with a reduced state vector and make significant time gains, while the forward/dynamics model still uses the full state in a simulator \cite{MuJoCo_2012}.

We propose and compare different methods to identify the relevant DoFs of the system. We first propose a \textit{naive} method, which considers a DoF relevant if and only if it appears in the cost function for a given task. While this in general gives a good heuristic, it can miss important DoFs that are not in the cost formulation (e.g., an object that is not considered in the cost, pushing another object that is in the cost), as well as include DoFs in the state vector that are not always relevant (e.g., an object that is in the cost formulation, but is not in a position to improve the cost given the current state and the nominal trajectory). Therefore, we also investigate methods that try to identify the relevant DoFs for the current state and around the current nominal trajectory. For this, we make use of the LQR gain matrix, $K$, which relates how changes in the current state should result in changes in current controls, for optimal behaviour. We propose two methods that use $K$: The first method directly uses the values in the columns of $K$ to identify the state elements that can induce a large change in controls for optimal behaviour, and therefore are considered relevant. The second method performs an SVD decomposition on $K$ to identify the principal axes and use them to identify the most important state elements.
We compare the three methods above to two baselines: A vanilla iLQR that uses the full state vector at all times, and finally a method that \textit{randomly} chooses subsets of the state vector during optimisation.

Our results show that the $K$-informed methods can identify the relevant DoFs of the task at different points during task execution (as can also be seen in Fig.~\ref{push_mcl_snapshots}), resulting in significantly lower optimisation times. Within an MPC loop, these lower optimisation times imply a lower policy lag, resulting in better MPC performance when compared with the baseline methods.

We make the following contributions in this paper\footnote{ Code and data: \href{https://github.com/DMackRus/iLQR_SVR}{here}}:
\begin{itemize}
    \item We propose a general MPC approach that changes the elements inside the state vector \textit{online} to minimise optimisation times, and therefore policy lag.
    \item We propose and compare different methods to identify the relevant DoFs inside the state vector given a task.
    \item We present our modifications to iLQR, to perform trajectory optimisation with a reduced state vector.
    \item We evaluate our methods on three high-dimensional non-prehensile manipulation tasks.
\end{itemize}
\section{Related work} \label{Related work}

Trajectory optimisation in robotics has traditionally been used for finding smooth, collision-free motion plans \cite{Kalakrishnan_2011_stomp,Ratliff_2009_chomp}. Recently, its applications have expanded to more complex manipulation and locomotion tasks, such as in-hand manipulation and grasping \cite{Charlesworth_dexterous_2021,ciocarlie_2007_dimensionality}, full-body manipulation \cite{Kurtz_2022_iLQR}, locomotion \cite{Posa_2014_optimisation}, and object manipulation in clutter \cite{Kitaev_2015,Agboh2018,Papallas_2020_optimisation}. 

One challenge of manipulation in clutter is physics uncertainty, where trajectories that are optimised in simulation often fail in the real-world due to unpredictable physical interactions. To address this challenge, Model Predictive Control (MPC) \cite{howell_2022_MJPC,williams2017model,pezzato2023sampling} is commonly used. MPC uses real-time model predictions to optimise control actions, allowing the system to adapt to changes and uncertainties as they occur. However, trajectory optimisation does not scale to high-dimensional spaces well. This is primarily due to the curse of dimensionality, which leads to long optimisation times. As the complexity of the task increases, so does the number of dimensions that need to be considered, making efficient optimisation and MPC increasingly difficult.

Various works explored different methods to speed up motion planning \cite{agboh2020pushing,saleem2020planning,Papallas_2020_optimisation}. For example, Saleem and Likhachev \cite{saleem2020planning} developed a planning method that uses a physics simulator only when necessary, relying on a cheaper geometric model for actions far from obstacles to reduce computation time. While these methods typically optimise a solution within several seconds, this duration poses challenges for real-time MPC.

Different methods for \textit{dimensionality reduction} in robotics have been explored. For instance, Rapidly-exploring Random Trees (RRT) struggles to find valid motion plans as the dimensionality of the search space increases. Zheng et al. \cite{zheng_2021_accelerating} consider reducing the dimensionality of the sample state space to reduce the time taken to find an optimal solution. Similarly, Vernaza et al. \cite{Vernaza_2011_dim_reduction,vernaza2012learning} consider locomotion problems for abstract robots that have a large number of DoFs that need to navigate around a cluttered space to some desired configurations whilst avoiding collisions. A different form of dimensionality reduction is considered by Hogan and Rodriguez \cite{Hogan_2016}, they formulate a direct transcription trajectory optimisation problem for a 2D push-slider system and consider the exponential nature of switching contact modes over a long trajectory horizon. To limit the exponential explosion, they limit the number of times the contact mode can switch over a trajectory to make the problem computationally tractable.

Manipulation of deformable objects is inherently a high-dimensional problem, often resulting in lengthy motion planning times. Wang et al. \cite{wang_2023_goal} consider the problem of manipulating 1D and 2D soft bodies (ropes and cloths) into goal configurations. To reduce computational costs, they computed a minimal set of key-points that their algorithm can manipulate, enabling efficient planning of folding tasks. Mahoney et al. \cite{mahoney_2010_deformable} considered reducing the dimensionality of a soft robot for computing collision free motion plans in tight environments, using principal component analysis. 

Notably, Sharma and Chakravorty \cite{Sharma_2023_RODiLQR} propose a reduced order iLQR implementation similar to ours, where the roll-outs are performed using the full system dynamics. They evaluate their methods for optimisation until convergence for given high-dimensional PDEs. Our work focuses on robotic manipulation problems where analytical formulations of the PDEs are not readily available. We also focus on dynamically changing the size of the state vector used during \textit{online} execution.

Ciorcarlie et al. \cite{ciocarlie_2007_dimensionality} consider reducing the dimensionality of grasping problems by restricting the number of grasps to a set of key ``eigengrasps''. They show that this reduced action space is still capable grasping a wide variety of complex objects. Jin et al. \cite{jin_2024_task} consider \textit{learning} a reduced-order hybrid model for complex three finger manipulation, enabling real-time closed loop MPC. Chen et al. \cite{Chen_2020_dim_reduction} consider the problem of learning a reduced order model for bipedal locomotion. In their work, they formulate a set of tasks for a bipedal robot to perform (walking on level ground, walking uphill etc) for which they want to compute a reduced order model offline that can be used for all their specified tasks. They manually specify a feature vector (a lower dimensional combination of features from the full-order model) and learn a set of linear weights for this feature vector offline by performing trajectory optimisation using the current reduced order model. The linear weights are adjusted via stochastic gradient descent.

These learning methods leverage prior distributions of the tasks to learn a reduced model. However, in robotic object manipulation, often the number and types of objects are unknown in advance, and each problem may require a different reduced model. We propose a method that can \textit{dynamically} change the system state over time during MPC execution. At any given point, we use information from the trajectory optimiser to identify the most relevant DoFs for the current task configuration and optimise in this reduced state space. Our approach does not assume any prior task distribution, and we evaluate its effectiveness for both \textit{rigid} and \textit{deformable} object manipulation in cluttered environments.

\section{Problem definition} \label{Problem definition}

We consider a discrete time dynamics formulation,

\begin{equation} \label{General dynamics}
     \textbf{x}_{t+1} = f(\textbf{x}_{t}, \textbf{u}_{t})
\end{equation}

\noindent where $\textbf{x}_t$ and $\textbf{u}_t$ are the state and control vectors respectively, at some time-step $t$ along a trajectory.

We have some running cost function;

\begin{equation} 
     l(\textbf{x}_{t}, \textbf{u}_{t}) = (\textbf{x}_{t} - \tilde{\textbf{x}}_{t})^{\mathsf{T}}W(\textbf{x}_{t} - \tilde{\textbf{x}}_{t}) + \textbf{u}_{t}^{\mathsf{T}}R\textbf{u}_{t}
     \label{eq:cost}
\end{equation}

\noindent where $W$ and $R$ are semi-positive definite cost weighting matrices and $\tilde{\textbf{x}}_{t}$ is the desired state at time-step $t$.

The total running cost of a trajectory, $J$, is the summation of all the running costs as well as the terminal cost function, when a control sequence 
$\textbf{U} \equiv (\textbf{u}_0, \textbf{u}_1, \cdots, \textbf{u}_{T-1} )$ is applied from some initial starting state $\textbf{x}_{0}$;

\begin{equation} \label{eq:total_running_cost}
     J(\textbf{x}_0, \textbf{U}) = l_{f}(\textbf{x}_{T}) + \sum_{t=0}^{T-1} l(\textbf{x}_{t}, \textbf{u}_{t})
\end{equation}

\noindent where $l_f$ is the terminal cost function of the same form as Eq.~\ref{eq:cost} but with different matrices $W_f$ and $R_f$. We also write the full trajectory state sequence that comes from rolling out the control sequence $\textbf{U}$ as $\textbf{X} \equiv (\textbf{x}_0, \textbf{x}_1, \cdots \textbf{x}_{T} )$.

The general trajectory optimisation problem is to compute an optimal sequence of controls  that minimise the total running cost of the trajectory, from some initial state:

\begin{equation} \label{eq:generalTrajOptProblem}
     \textbf{U}^*(\textbf{x}_0) = \arg\min_{\textbf{U}} J(\textbf{x}_0, \textbf{U})
\end{equation}

We are interested in using trajectory optimisation within a model predictive control (MPC) framework. During each MPC iteration, the optimisation problem from Eq. \ref{eq:generalTrajOptProblem} is solved, and the optimised controls are executed on the \textit{real}\footnote{In this work, we use a simulator to also represent the \textit{real} system.} system, while another round of optimisation is initiated with the current state of the system. Therefore, the \textit{control frequency} is determined by how quickly Eq.~\ref{eq:generalTrajOptProblem} is solved, and it has a significant effect on the performance of the MPC scheme. 

Let $\underline{\textbf{x}}_t$ represent the real system state during MPC execution at time $t$, and suppose the MPC executes $Y$ controls before it stops\footnote{We stop the controller if a \textit{Task timeout} is reached, or if a success condition is achieved.}. We use $\texttt{MPC\_Cost}$ to quantify the MPC performance:
\begin{equation}\label{eq:mpc_cost}
\texttt{MPC\_Cost} =  l_{f}(\underline{\textbf{x}}_{Y}) + \sum_{t=0}^{Y-1} l(\underline{\textbf{x}}_{t}, \textbf{u}_{t})
\end{equation}
The expression above is similar to Eq.~\ref{eq:total_running_cost} and uses the same cost functions $l$ and $l_f$. The difference is that, while Eq.~\ref{eq:total_running_cost} is evaluated over the states $\textbf{x}_t$ as predicted by the dynamics model, Eq.~\ref{eq:mpc_cost} is evaluated over the real states achieved by the system. Furthermore, while Eq.~\ref{eq:total_running_cost} sums over the optimisation horizon $T$, Eq.~\ref{eq:mpc_cost} sums over the complete duration of the execution, $Y$.

In this work, our aim is to improve the $\texttt{MPC\_Cost}$ by solving Eq.~\ref{eq:generalTrajOptProblem} faster, and therefore achieving a higher control frequency during MPC.

\subsection{Definitions}

We consider manipulation tasks consisting of a robot with $q$ joints and multiple objects. We have $N_O$ rigid objects and $N_S$ deformable/soft objects.  Rigid objects simply have 6 degrees of freedom representing their pose. Deformable object $i$ consists of $N_{S_i}$ particles, and has three DoFs $\{x, y, z\}$ per particle. We use $\mathcal{F}$ to refer to the \textit{set} of all DoFs in the system. Therefore, 

\begin{equation} \label{dofs_F}
    |\mathcal{F}| = q + 6N_O + 3\sum_{i=1}^{N_S}N_{S_i}
\end{equation}

In this work we are interested in discovering and using a reduced set of DoFs, $\mathcal{C} \subseteq \mathcal{F}$. We will use superscript notation when referring to the reduced versions of vectors or matrices of our system. For example, $\textbf{x}^{\mathcal{C}}_t$ refers to the reduced state vector which only includes elements corresponding to DoFs in $\mathcal{C}$. We do \textit{not} use a superscript when the full set of DoFs, $\mathcal{F}$, is used.

We consider both positional and velocity elements for each DoF inside our state vector. Therefore, $\textbf{x}_t \in \mathbb{R}^{2|\mathcal{F}|}$ whereas $\textbf{x}^{\mathcal{C}}_t \in \mathbb{R}^{2|\mathcal{C}|}$. The control vector  has size $m$, $\textbf{u}_t \in \mathbb{R}^{m}$, and does not change size.

We denote the set of DoFs \textit{not} currently in our reduced set of DoFs as $\mathcal{L}$, i.e.,  $\mathcal{L} = \mathcal{F} \setminus \mathcal{C}$.
\section{Method}
\begin{algorithm}[b!] 
\caption{MPC with state vector reduction (asynchronous)} \label{General_algorithm}  
	\begin{algorithmic}[1] 
        \State{$\mathcal{C} \gets $ \textproc{InitialiseSubset($\mathcal{F}$)}}
        \State{$\mathcal{L} \gets \mathcal{F} \setminus \mathcal{C}$}
        \State{$\textbf{U} \gets $ \textproc{InitialiseControls()}}

        \Statex{\textbf{Optimiser Asynchronous (Planning)}}
            \Indent
            \While{task \textbf{not} complete}
                \State{$\textbf{x}_0 \gets $ \textproc{GetCurrentState()} \label{algline:get_state}}
    
                \State{$\texttt{dofs\_to\_add} \gets$ \textproc{IdentifyDoFsToAdd($\mathcal{L}$)}} \label{algline:adddofs} 
                \State{$\mathcal{C} \gets \mathcal{C} \cup \texttt{dofs\_to\_add} $}
                \State{$\mathcal{L} \gets \mathcal{L} \setminus \texttt{dofs\_to\_add} $}

                \State{$\textbf{U}, \textbf{K}^\mathcal{C} \gets$ \textproc{Optimise}($\textbf{x}_0, \textbf{U}, \mathcal{C}$)} \label{algline:optimise}
    
                \State{$\texttt{dofs\_to\_remove} \gets$ \textproc{IdentifyDoFsToRemove($\textbf{K}^\mathcal{C}$)}} \label{algline:removedofs}
                \State{$\mathcal{C} \gets \mathcal{C} \setminus \texttt{dofs\_to\_remove} $}
                \State{$\mathcal{L} \gets \mathcal{L} \cup \texttt{dofs\_to\_remove} $}
            \EndWhile
            \EndIndent
        \Statex{\textbf{Agent Asynchronous  (Execution)}}
        \Indent
        \While{task \textbf{not} complete}
            \State{Execute first control from $\textbf{U}$}
            \State{Remove first control from $\textbf{U}$}
            \State{Pad $\textbf{U}$ with the last control}
        \EndWhile
        \EndIndent
	\end{algorithmic} 
\end{algorithm}

A general asynchronous MPC scheme (similar to Howell et al. \cite{howell_2022_MJPC}) can be seen in Alg.~\ref{General_algorithm}. An \textit{agent} continuously executes an optimised control sequence $\textbf{U}$. The optimiser continuously queries the current state from the agent (line \ref{algline:get_state}) and optimises the control sequence $\textbf{U}$ (line \ref{algline:optimise}). Depending on how long the optimiser takes to optimise the control sequence $\textbf{U}$\footnote{In MPC, we only perform one iteration of optimisation, not optimisation until convergence.}, the agent moves beyond the state that the optimiser is initialized with. The consequence of this is that the control sequence becomes less useful the longer optimisation takes; this is often referred to as \textit{policy lag}.

In this work, we augment this general MPC formulation to dynamically change the size of the state vector currently being used in line \ref{algline:optimise}, to reduce optimisation times, thus decreasing policy lag. We initialise our reduced set of DoFs in line 1: in this work we simply set $\mathcal{C} = \mathcal{F}$, but if a better heuristic is available that can be used instead. Before every optimisation call, we identify a number of unused DoFs in $\mathcal{L}$ to reintroduce to our reduced set of DoFs $\mathcal{C}$ on line \ref{algline:adddofs}. We then optimise a control sequence using this reduced set of DoFs and after computing a control sequence, we identify a set of DoFs that were  unimportant to the previous trajectory optimisation problem on line \ref{algline:removedofs}. These DoFs are then removed from our reduced set of DoFs.


There are three important components of this general approach. Firstly, we need a method of optimising a trajectory using only a reduced set of DoFs (Alg.~\ref{General_algorithm}, line \ref{algline:optimise}). This is explained in section \ref{optimise} where we formulate performing iLQR \cite{Li_2004_iLQR,Tassa_2012_iLQR} on our subset of reduced DoFs. Secondly, a method of determining which DoFs can be removed is required (Alg.~\ref{General_algorithm}, line \ref{algline:removedofs}). 
In Sec.~\ref{reduce dim} we outline different methods we propose to identify the DoFs to be removed. Finally, a method for reintroducing DoFs into our reduced set of DoFs is required (Alg.~\ref{General_algorithm}, line \ref{algline:adddofs}). In this work, we propose a simple random sampling strategy from the unused DoFs $\mathcal{L}$. It would be trivial to improve on the method for reintroducing DoFs into the system if specific information about the task and current trajectory was exploited. However, the purpose of this work was to suggest an abstract method for online state vector reduction that is task agnostic, as such we did not purse this line of work.

\subsection{Optimise}  \label{optimise}

In this section we outline how we augment the iLQR \cite{Li_2004_iLQR,Tassa_2012_iLQR} algorithm to operate on our reduced set of DoFs $\mathcal{C}$. The high level overview is that we perform derivative computation and backwards pass calculations only for our reduced set of DoFs. We use the MuJoCo simulator to model the full system. When we perform forwards-rollouts, the full set of DoFs is updated using MuJoCo, as well as computing the total running cost of the trajectory. We name our iLQR adaption \textit{iLQR with state vector reduction} (iLQR-SVR).

iLQR-SVR uses a first order approximation of the system dynamics where we write Eq. \ref{General dynamics} as Eq. \ref{First order dynamics}.

\begin{equation} \label{First order dynamics}
    \textbf{x}^{\mathcal{C}}_{t+1} = A_{t}^{\mathcal{C}}\textbf{x}^{\mathcal{C}}_{t} + B_{t}^{\mathcal{C}}\textbf{u}_t
\end{equation}

\noindent where $A^{\mathcal{C}}_{t} = \delta f(\textbf{x}^{\mathcal{C}}_{t}, \textbf{u}_t) / \delta \textbf{x}^{\mathcal{C}}_t$ and $B_{t}^{\mathcal{C}} = \delta f(\textbf{x}^{\mathcal{C}}_{t}, \textbf{u}_t) / \delta \textbf{u}_t$. iLQR also requires a first and second order approximation of the cost derivatives with respect to the state and control vector $(l^{\mathcal{C}}_\textbf{x}, l^{\mathcal{C}}_{\textbf{xx}}, l_\textbf{u}, l_{\textbf{uu}})$.

The computation of the dynamics derivatives $\textbf{A} \equiv \{ A_0, A_1, \cdots A_{T-1} \}$ and $\textbf{B}  \equiv \{ B_0, B_1, \cdots B_{T-1} \}$ is often the bottleneck in gradient-based trajectory optimisation. These derivatives usually need to be computed via computationally costly \textit{finite-differencing}. Finite-differencing requires evaluating the system dynamics for every DoF in the system as well as any control inputs. By only computing dynamics derivatives for our reduced state ($\textbf{A}^\mathcal{C}$ and $\textbf{B}^\mathcal{C}$) we reduce the number of dynamics evaluations to $2|\mathcal{C}| + m$ instead of $2|\mathcal{F}| + m$. 

Using these approximations, optimal control modifications can be computed recursively using the dynamic programming principle \cite{Li_2004_iLQR,Tassa_2012_iLQR}. This step is colloquially referred to as the \textit{backwards pass}. and works by propagating the value function $V$ from the end of the trajectory to the beginning by computing equations 8 - 10 from $t=T$ to $t=0$. We augment the following equations to operate on our reduced set of DoFs:

\begin{subequations}
\begin{align}
Q_\textbf{x}^\mathcal{C} &= l_\textbf{x}^\mathcal{C} + (A^{\mathcal{C}})^\mathsf{T}V^{'\mathcal{C}}_{\textbf{x}}\\
Q_\textbf{u} &= l_\textbf{u} + (B^{\mathcal{C}})^\mathsf{T}V^{'\mathcal{C}}_{\textbf{x}}\\
Q_\textbf{xx}^\mathcal{C} &= l_\textbf{xx}^\mathcal{C} + (A^{\mathcal{C}})^\mathsf{T}V^{'\mathcal{C}}_{\textbf{xx}}A^{\mathcal{C}}\\
Q_\textbf{uu} &= l_\textbf{uu} + (B^{\mathcal{C}})^\mathsf{T}V^{'\mathcal{C}}_{\textbf{xx}}B^{\mathcal{C}}\\
Q_\textbf{ux}^\mathcal{C} &= l_\textbf{ux}^\mathcal{C} + (B^{\mathcal{C}})^\mathsf{T}V^{'\mathcal{C}}_{\textbf{xx}}A^{\mathcal{C}}
\end{align}
\end{subequations}

\noindent At every time-step, these Q matrices can be used to compute an open-loop feedback term $k$ as well as a closed-loop state feedback gain $K$.

\begin{subequations}
\begin{align}
k_t &= -Q_{\textbf{uu}}^{-1}Q_\textbf{u}\\
K^\mathcal{C}_t &= -Q_{\textbf{uu}}^{-1}Q_\textbf{ux}^\mathcal{C}
\end{align}
\end{subequations}

\noindent Finally the value function needs to be updated.

\begin{subequations}
\begin{align}
V_\textbf{x}^\mathcal{C} &= Q_\textbf{x}^{\mathcal{C}} - Q_\textbf{u}Q_{\textbf{uu}}^{-1}Q_{\textbf{ux}}^\mathcal{C}\\
V_{\textbf{xx}}^\mathcal{C} &= Q_\textbf{xx}^{\mathcal{C}} - Q_\textbf{u}Q_{\textbf{uu}}^{-1}Q_{\textbf{ux}}^\mathcal{C}
\end{align}
\end{subequations}

\noindent iLQR-SVR performs a forwards roll-out using the MuJoCo model, i.e., using the full non-linear system dynamics. We update the state vector for the full set of DoFs whilst only computing control modifications based on our reduced set of DoFs, as shown by Eq. \ref{Feedback gain}.

\begin{equation}
    \hat{\textbf{x}}_0 = \textbf{x}_0
\end{equation}
\begin{equation} \label{Feedback gain}
    \hat{\textbf{u}}_t = \textbf{u}_t + \alpha k_t + K^{\mathcal{C}}_{t}(\hat{\textbf{x}}^{\mathcal{C}}_t - \textbf{x}^{\mathcal{C}}_t)
\end{equation}
\begin{equation}
    \hat{\textbf{x}}_{t+1} = f(\hat{\textbf{x}}_{t}, \hat{\textbf{u}}_{t})
\end{equation}

\noindent Above, $\hat{\textbf{x}}_{t}, \hat{\textbf{u}}_{t}$ denotes the new computed trajectory states and controls and $\alpha$ is a line-search parameter between 0 and 1. If a lower cost trajectory is found, the nominal state and control trajectory is updated from the roll-out.

\remark{If a DoF is removed from $\mathcal{C}$, that DoF will still incur a cost when computing the running cost of a trajectory, i.e the cost function \textbf{always} operates over the full set of DoFs $\mathcal{F}$.}

\subsection{Reducing dimensionality}    \label{reduce dim}

In this section, we outline how we identify DoFs to be removed from the system state. The core idea of our methods is leveraging information from the LQR gain matrices, $K$, which in our case, are already computed and returned by the iLQR algorithm, as the state-feedback gain matrices. In case another optimisation algorithm is used, then a linearised LQR approximation could be applied about the nominal trajectory, to compute a closed-loop state-feedback gain around it.

The $\textbf{K} \equiv \{ K_1, K_2\cdots, K_{T-1}\}$ matrices compute a closed-loop control modification to add to the nominal control vector when performing a roll-out using full non-linear dynamics. This is achieved by calculating the difference in the current state along the new trajectory against the nominal trajectory for which derivatives were computed. This difference is then multiplied by a linear gain (i.e the $K$) matrix. This is shown explicitly for a single time-step in Eq. \ref{Matrix feedback gain}. 

\begin{equation} \label{Matrix feedback gain}
\begin{bmatrix}
    u(0) \\
    u(1) \\
    \vdots\\
    u(m)
\end{bmatrix}
=
\begin{bmatrix}
    K^\mathcal{C}(0,0) & K^\mathcal{C}(0,1) & \cdots & K^\mathcal{C}(0, 2|\mathcal{C}|)\\
    K^\mathcal{C}(1,0) & K^\mathcal{C}(1,1) & \cdots & K^\mathcal{C}(1, 2|\mathcal{C}|)\\
    \vdots  & \vdots & \ddots & \vdots\\
    K^\mathcal{C}(m,0) & K^\mathcal{C}(m,1) & \cdots & K^\mathcal{C}(m, 2|\mathcal{C}|)\\
\end{bmatrix}
\begin{bmatrix}
    \hat{x}^\mathcal{C}(0) - x^\mathcal{C}(0) \\
    \hat{x}^\mathcal{C}(1) - x^\mathcal{C}(1) \\
    \vdots \\
    \hat{x}^\mathcal{C}(2|\mathcal{C}|) - x^\mathcal{C}(2|\mathcal{C}|)
\end{bmatrix}
\end{equation}

\noindent We use notation $K(.,.)$ to denote indexing inside the matrix. The \textbf{columns} inside the $K$ matrix correspond to computing entire control vector modifications based on the deviation of a specific DoF from the nominal. We reason that, DoFs that have small gain values associated with them over the entire trajectory are not important to the overall trajectory optimisation problem and can be ignored. This is because even if that DoF had a large deviation from its nominal position, it would have a minimal impact on the newly computed control.

We consider two approaches of leveraging this information, a) simply summing up the relevant values inside these matrices, b) performing singular value decomposition (SVD) to try find the most relevant DoFs. Both methods compute \texttt{dof\_importance} values for every DoF in $\mathcal{C}$, we then simply use a threshold parameter $\rho$ to determine which DoFs should remain in the $\mathcal{C}$ and which ones should be removed.

\subsubsection{Summing}

The summing method simply leverages the values inside the $K$ matrix for each DoF and how much of an impact they have on the control vector modification. If the values inside the $K$ matrix are large for a particular DoF over the entire optimisation horizon, then that DoF is likely important to the trajectory optimisation problem currently. We compute a $\texttt{dof\_importance}$ value for each DoF using the following formula:

\begin{equation} \label{dof importance}
\texttt{dof\_importance}[j] = \sum_{t=0}^{T}\sum_{p=0}^{m}\Big(K^\mathcal{C}_{t}(p, j) + K^\mathcal{C}_{t}(p, j + |\mathcal{C}|)\Big) \Big/ T
\end{equation}

\noindent We sum over the number of controls in the control vector $m$ as well as summing over all matrices over the optimisation horizon. Finally we sum both the columns corresponding to the positional and velocity element for the DoF $j$.

\subsubsection{SVD}

The SVD method performs singular value decomposition (SVD) on the state feedback gain matrices:

\begin{equation}
    K_t = U_t\Sigma_t V_t^\mathsf{T}
\end{equation}

\noindent where $K$ is our state feedback gain matrix of size $m \times 2|\mathcal{C}|$, $U$ is an orthogonal $m \times m$ matrix, $\Sigma$ is a positive diagonal matrix  of size $m \times 2|\mathcal{C}|$ and finally $V$ is also an orthogonal matrix of size $2|\mathcal{C}| \times 2|\mathcal{C}|$.

The diagonal values of $\Sigma$ are the singular values $\sigma_0 > \sigma_1 > \ldots \sigma_m$. We use the first $g$ singular values (we used $g=3$ in this work) and singular vectors in $V$ to extract the dominant DoFs in the state vector. Specifically, we calculate \texttt{dof\_importance} values for all DoFs in our current reduced set of DoFs, as:

\begin{equation} \label{dof_importance_svd}
\texttt{dof\_importance}[j] = \sum^T_{t=0}\sum^g_{n=0}\Big(\big(V_t(j, n) + V_t(j + |\mathcal{C}|, n)\big)\sigma_n\Big) \Big/ T
\end{equation}

\noindent We use $V(.,.)$ to denote indexing inside the matrix. Notably, for both methods of computing \texttt{dof\_importance} values, we divide by the optimisation horizon so that $\rho$ does not need to change dependant on the optimisation horizon used.

In both of these methods, we use the \texttt{dof\_importance} values to determine which DoFs to remove for Alg. \ref{General_algorithm} line \ref{algline:removedofs}. Any DoFs that have an importance value that is below $\rho$ are removed from the current reduced set of DoFs $\mathcal{C}$.  One exception to this is that  the robot DoFs are not allowed by any of the methods to be removed; i.e., the robot DoFs always appear in the state vector.




\section{Results} \label{Results}

\begin{figure}[b] \label{push_soft_snapshots}
\includegraphics[width=\textwidth]{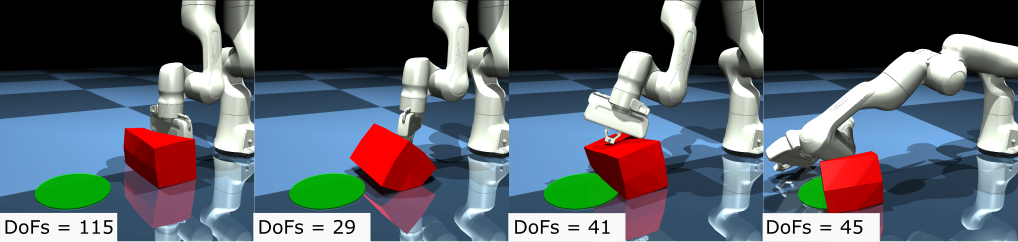}
\caption{A sequence of snapshots showing an example trajectory for the \textbf{soft} task. The objective is to push the red deformable object to the green flat circle on the floor. The full number of DoFs in this system is 115.}
\label{fig:soft}
\end{figure}

\begin{figure}[t] \label{push_soft_rigid_snapshots}
\includegraphics[width=\textwidth]{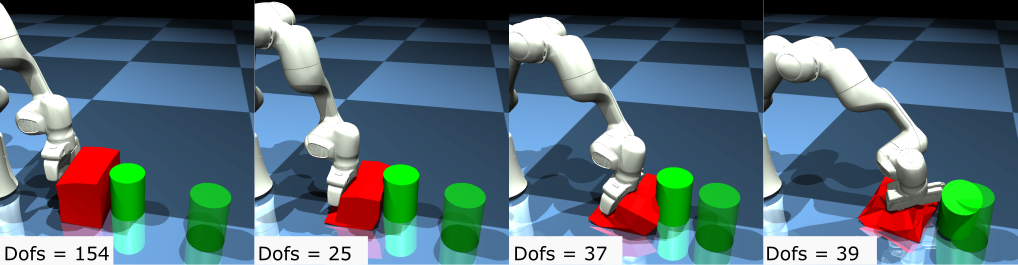}
\caption{A sequence of snapshots showing an example trajectory for the \textbf{soft rigid} task. The objective is to move the green cylinder to the goal region (the transparent cylindrical region) with a high-dimensional soft body being placed between the robot end-effector and the green cylinder. The full number of DoFs in this system is 154.} 
\label{fig:soft_rigid}
\end{figure}

We perform simulation experiments under an asynchronous MPC framework as described in Alg. \ref{General_algorithm}. We provide results for a variety of methods. We have two baseline methods:
\begin{itemize}
    \item \textit{iLQR-Baseline}: A baseline method, where we perform iLQR as normal on the full set of DoFs in the system.
    \item \textit{iLQR-Rand}: A baseline method where we perform iLQR on a random reduced set of DoFs. Instead of using any intelligent method to determine which DoFs, we instead randomly sample $\theta$ DoFs to use (in addition to the robot DoFs) in every optimisation iteration.
\end{itemize}

\noindent We have three methods that we evaluate against the baselines: 

\begin{itemize}
    \item \textit{iLQR-Naive}: This method reduces the set of DoFs in our reduced set to only the DoFs that are directly considered in the cost function for the task.
    \item \textit{iLQR-SVR-SVD}: Our method of dynamically changing the number of DoFs as described in Sec.~\ref{reduce dim}-SVD.
    \item \textit{iLQR-SVR-Sum}: Our method of dynamically changing the number of DoFs as described in Sec.~\ref{reduce dim}-Summing.
\end{itemize}

All experiments were performed on a 16-core 11th Gen Intel(R) Core(TM) i7-11850H @ 2.50G with 32GB of RAM.

\subsection{Task definition}
We selected three high-dimensional non-prehensile manipulation tasks to provide experiments on, the main idea behind these tasks was to create tasks that were not too mechanically challenging but were high dimensional to see if our methods are capable of determining what DoFs are needed to solve the tasks efficiently. In some of the tasks that have very large state spaces, we were required to slow down the simulated agent thread by some factor to maintain some level of performance. All tasks used a Franka Panda robotic arm with 7 actuated robotic joints (not considering the grippers in the control vector).

We outline the general description of the three tasks as well as their task hyper-parameters below. $T$ is the optimisation horizon, $\Delta t$ was the model time-step. $Y$ is the task timeout which is the number of time-steps in the agent thread until the task was finished. Finally, slowdown factor was how many times slower the agent thread was compared to the model time-step.

\textbf{Clutter task:} The aim of this task is to push a green cylinder to a target goal region (shown by a green silhouette) whilst minimally disturbing a set of distractor objects (example in Fig. \ref{push_mcl_snapshots}). The task hyper-parameters were as follows; $T = 80$, $\Delta t = 0.004$, $Y = 2000$, Slowdown factor = $1$.

\textbf{Soft task:} The aim of this task is to push a soft body.
This task aimed to push a high dimensional red soft body to a goal location (example in Fig. \ref{fig:soft}). The task hyper-parameters were as follows; $T = 50$, $\Delta t = 0.004$, $Y = 1000$, Slowdown factor = $3$.

\textbf{Soft rigid task:} This task aimed to push a green cylinder to a target location (green silhouette) but there is a  high dimensional soft body in between the robot end-effector and the goal object. This means that the optimiser needs to reason about the dynamics between the soft body and the rigid body to achieve the task (example in Fig. \ref{fig:soft_rigid}). The task hyper-parameters were as follows; $T = 100$, $\Delta t = 0.004$, $Y = 1000$, Slowdown factor = $5$.

\subsection{Asynchronous MPC results}

\begin{table}[t]
\centering
\caption{Results of asynchronous MPC for three manipulation tasks. The values in the table are averaged over 100 trials for the clutter task and 20 trials for the soft and soft rigid task. The first value is the mean and the second is the 90\% confidence interval. For the three first methods, the number of DoFs is a preset parameter, while the rest of the methods adjust the number of DoFs dynamically.} \label{results_table}
\begin{tabular}{|>{\centering\arraybackslash}m{2.5 cm}c|c|c|c|}  
\hline
\multicolumn{2}{|c|}{\textbf{Method / Task}} & \textbf{Clutter} & \textbf{Soft} & \textbf{Soft rigid} \\
\hline
\multirow{3}{*}{iLQR-Baseline} & MPC cost & $1.00\pm0.034$& $1.00\pm0.080$& $1.00\pm0.096$\\
 & Opt time (ms)& $485.71\pm5.16$& $620.97\pm7.35$& $4322.39\pm366.65$\\
 & Num DoFs (preset)& $55$& $115$& $154$\\
\hline
\multirow{2}{*}{iLQR-Rand} & MPC cost & $0.94\pm0.033$& $1.14\pm0.104$& $0.74\pm0.071$\\
 & Opt time (ms)& $148.91\pm1.69$& $79.12\pm4.13$& $881.54\pm7.67$\\
 $\theta = 5$ & Num DoFs (preset)& $12$ & $12$ & $12$\\
\hline
\multirow{3}{*}{iLQR-Naive} & MPC cost & $0.89\pm0.035$& $0.92\pm0.067$& \boldmath{$0.38\pm0.066$}\\
 & Opt time (ms)& $219.07\pm2.11$& $395.52\pm7.71$& $583.06\pm85.72$\\
 & Num DoFs (preset) & $23$& $79$& $9$\\
\hline
\multirow{2}{*}{iLQR-SVR-SVD}& MPC cost &\boldmath{$0.86\pm 0.033$}& \boldmath{$0.90 \pm 0.065$}& $0.88 \pm 0.071$\\
 & Opt time (ms)& $335.16 \pm 13.61$& $367.18 \pm 49.60$& $2320.80 \pm 392.91$\\
 $\theta=$10, $\rho=$1& Num DoFs& $31.85 \pm 1.12$& $72.55 \pm 9.37$& $85.19 \pm 18.32$\\
\hline

\multirow{2}{*}{iLQR-SVR-SVD}& MPC cost & $0.91 \pm 0.036$& $0.91 \pm 0.069$& $0.66 \pm 0.063$\\
 & Opt time (ms)& $237.79 \pm 5.76$& $116.20 \pm 4.73$& $1051.20\pm 32.55$\\
 $\theta=$10, $\rho=$500& Num DoFs& $21.06 \pm 0.33$& $23.32 \pm 0.77$& $24.72 \pm 0.17$\\
\hline
\multirow{2}{*}{iLQR-SVR-SVD} & MPC cost & $0.91 \pm 0.036$& $0.90 \pm 0.061$& $0.89 \pm 0.068$\\
 & Opt time (ms)& $271.21 \pm 11.41$& $251.90 \pm 38.30$& $1296.29 \pm 184.13$\\
 $\theta=$5, $\rho=$1& Num DoFs& $24.38 \pm 1.00$& $49.59 \pm 7.34$& $38.69 \pm 8.66$\\
\hline
\multirow{2}{*}{iLQR-SVR-Sum}& MPC cost & $0.86\pm0.032$& $0.94\pm0.073$& $0.77\pm0.074$\\
 & Opt time (ms)& $287.05\pm10.91$& $136.18\pm10.92$& $1130.83 \pm 54.68$\\
 $\theta=$10, $\rho=$1& Num DoFs& $29.82\pm1.01$& $26.68\pm2.08$& $29.10\pm2.92$\\
\hline
\multirow{2}{*}{iLQR-SVR-Sum}& MPC cost & $0.87\pm0.029$& $1.00\pm0.081$& $0.70\pm0.071$\\
 & Opt time (ms)& $192.59\pm2.08$& $101.15\pm4.73$& $997.71\pm32.55$\\
 $\theta=$10, $\rho=$500& Num DoFs& $18.69\pm0.10$& $18.20\pm0.11$& $24.09\pm0.46$\\
\hline
\multirow{2}{*}{iLQR-SVR-Sum} & MPC cost & $0.94\pm0.042$& $0.97\pm0.079$& $0.75\pm0.066$\\
 & Opt time (ms)& $218.45\pm8.38$& $114.74\pm13.76$& $935.45\pm37.03$\\
 $\theta=$5, $\rho=$1& Num DoFs& $21.91\pm0.86$& $19.94\pm1.87$& $20.21\pm0.91$\\
\hline
\end{tabular}
\end{table}

In this section we evaluate the performance of our proposed methods on the three outlined tasks. Earlier, in Eq. \ref{eq:mpc_cost} we defined the \texttt{MPC\_cost} as the cost of the \textit{actual} trajectory executed in the real system. We use this as our metric of success. We show that our methods are capable of reducing the \texttt{MPC\_cost} compared to the baselines, even though they are optimising only on a reduced state vector. This is because that whilst individual optimisation performance will be somewhat lower when only considering a subset of DoFs, the significant optimisation time savings reduce the detrimental effects of \textit{policy lag}.

Table~\ref{results_table} shows the results from our asynchronous MPC experiments. We performed 100 runs for the clutter task and 20 runs for the soft and soft rigid tasks. The first value in the table are the mean values and the second values are 90\% confidence intervals. It should be noted that we removed outliers for the \textbf{clutter} task to prevent individual data points from disproportionately affecting the results. In this task, due to cylinders being used as the goal and distractor objects, in certain occasions if a cylinder was toppled with some large velocity, it would roll into the distance away from its goal position until the task timeout. The result of this would be a disproportionately high cost, in situations where this occurred we removed these outliers. Finally, we normalise all the cost values with respect to the \textit{iLQR Baseline} method.

The results show that, for all three tasks, our methods (\textit{iLQR-Naive}, \textit{iLQR-SVR-SVD} and \textit{iLQR-SVR-Sum}) are capable of achieving a lower $\texttt{MPC\_Cost}$ than the \textit{iLQR-Baseline}. \textit{iLQR-Naive} on average reduces the \texttt{MPC cost} by 27\% over all three tasks. Choosing the best parameterisation of our dynamic methods, they manage to lower the MPC cost by 14\%. Importantly, we show that simply limiting the size of the state vector does not achieve the same level of performance as our intelligent methods as \textit{iLQR-Rand} only reduces the \texttt{MPC cost} by 6\%.

These results are skewed slightly by the \textbf{soft rigid} task, where the \textit{iLQR-Naive} method significantly outperformed all other methods. If we only consider the \textbf{clutter} and \textbf{soft} tasks then \textit{iLQR-Naive} reduces the \texttt{MPC\_cost} by 8.5\%, \textit{iLQR-SVR} by 12\%, whereas \textit{iLQR-Rand} actually increase the \texttt{MPC\_cost} by 4\%.

This implies that our K-informed methods are more capable in these two tasks. This makes sense as the Naive methods always include DoFs in the state vector that are not necessarily important. For example, in the \textbf{clutter} task, when some objects are not involved in contact with the goal object / robot directly or indirectly, considering their dynamical properties adds nothing of value to the trajectory optimisation solution. Our K-informed methods are capable of deducing this automatically.

\begin{figure}[t]
    \centering
    \scriptsize
    \def\svgwidth{\linewidth}
    \scalebox{1.0}{\input{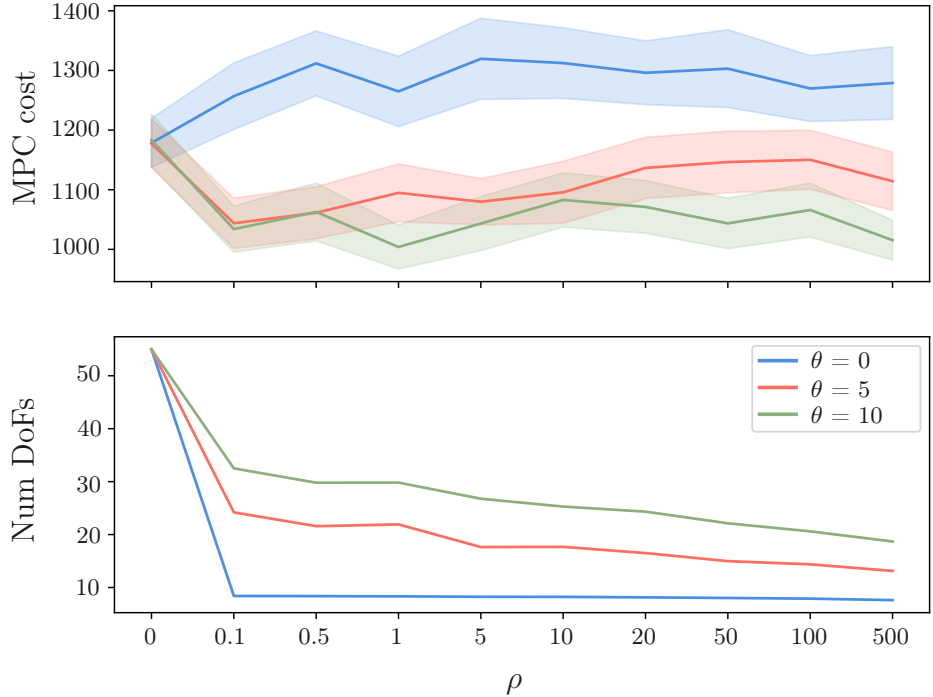}}
    \caption{Top plot shows the \texttt{MPC cost} averaged over 100 runs for the clutter task. The lightly shaded area shows the 90\% confidence interval range. Bottom plot shows the average number of DoFs in our state vector. Three different parameterisations of $\theta$ were used with scaling values for $\rho$.}
    \label{scaling_rho}
\end{figure}

It is important to recognise that we only show three parameterisations of $\theta$ and $\rho$ in these results and we keep them the same for all three tasks to test the generalisability of our methods. Better results could have been achieved if these values were tuned to each task specifically.

We also experimented with the effects of modifying the values for $\theta$ and $\rho$ for the clutter task, the results of this can be seen in Fig.~\ref{scaling_rho}. We experiment with three different values for $\theta$ and scale the value of $\rho$ between 0 and 500 (please note the values are not spaced equally). Firstly, when $\rho$ is set to zero, \textit{iLQR-SVR} operates identically to \textit{iLQR-Baseline}. This can be seen clearly in Fig.~\ref{scaling_rho} as all three data points have nearly identical \texttt{MPC Costs}. As the value for $\rho$ is increased, it makes our methods ``pickier'' at what DoFs remain in the state vector. This can clearly be seen in the bottom plots where the average number of DoFs in the state vector decreases as $\rho$ increases.

When $\theta$ is set to zero, no DoFs are ever re-introduced to our reduced set of DoFs. This means that once a DoF has been removed it never has the opportunity to be reconsidered in optimisation. This is clearly a bad strategy as it negates the core idea of our methodology which is that DoFs importance can change over time during a task, noticeably in Fig. \ref{scaling_rho}, this value of $\theta$ performs noticeably worse compared to the other two parameterisations of $\theta$.

The optimal parameterisation for $\rho$ would be some moderate value, as when $\rho$ approaches zero \textit{iLQR-SVR} performance should tend towards \textit{iLQR-Baseline} and when $\rho$ tends towards large values, \textit{iLQR-SVR} performance tend towards \textit{iLQR-Rand}. 

\section{Discussion} \label{Discussion}

In this work, we have made several contributions, firstly we have outlined a general method for changing the size of a state vector to be used in MPC \textit{online}. Secondly we have outlined an intelligent method of reducing the number of DoFs considered in trajectory optimisation and shown that our methods can increase performance when used in an asynchronous MPC format.


An unfortunate consequence of using a thresholding method as we have outlined in this work to decide whether DoFs should or should not be included in the state vector is that the threshold can sometimes be hard to tune and can be task dependant. More work is required to determine a better method to tune this parameter automatically. One such method that could perhaps be more robust would be to decide on a \textit{desired} state vector size prior to performing MPC. Instead of removing DoFs if they fall below some importance threshold, we could instead retain the most important DoFs up to our desired state vector size.

In this work we only consider discrete methods of reducing the size of the state vector, i.e a DoF is either considered completely or it is not. A promising line of future work would be to consider more conventional methods of dimensionality reduction, by combing DoFs into some lower dimensional space that could be discovered using principal component analysis.

\section*{Acknowledgments}
This research has received funding from the UK Engineering and Physical Sciences Research Council under the grants EP/V052659/1 and 2596473. For the purpose of open access, the authors have applied a Creative Commons Attribution (CC BY) license to any Author Accepted Manuscript version arising.


\end{document}